\documentclass[letterpaper]{article} 
\usepackage{aaai25}  
\usepackage{times}  
\usepackage{helvet}  
\usepackage{courier}  
\usepackage[hyphens]{url}  
\usepackage{graphicx} 
\usepackage{natbib}  
\usepackage{caption} 
\usepackage{booktabs}

\usepackage{algorithm}
\usepackage{amsmath}
\usepackage{algpseudocode}
\usepackage{newfloat}
\usepackage{listings}
\DeclareCaptionStyle{ruled}{labelfont=normalfont,labelsep=colon,strut=off} 
\floatstyle{ruled}
\newfloat{listing}{tb}{lst}{}
\floatname{listing}{Listing}
\title{LLM-based Online Prediction of Time-varying Graph Signals}
\author{
    Dayu Qin\equalcontrib, 
    Yi Yan\equalcontrib,
    Ercan Engin Kuruoglu\thanks{Corresponding author. Email: kuruoglu@sz.tsinghua.edu.cn. This work is supported by Tsinghua Shenzhen International Graduate School Start-up fund under Grant QD2022024C, Shenzhen Science and Technology Innovation Commission under Grant JCYJ20220530143002005 and Shenzhen Ubiquitous Data Enabling Key Lab under Grant ZDSYS20220527171406015.}    
}
\affiliations{
    Tsinghua-Berkeley Shenzhen Institute, Tsinghua Shenzhen International Graduate School, Tsinghua University
}

\begin{document}

\maketitle

\begin{abstract}
In this paper, we propose a novel framework that leverages large language models (LLMs) for predicting missing values in time-varying graph signals by exploiting spatial and temporal smoothness. We leverage the power of LLM to achieve a message-passing scheme. For each missing node, its neighbors and previous estimates are fed into and processed by LLM to infer the missing observations. Tested on the task of the online prediction of wind-speed graph signals, our model outperforms online graph filtering algorithms in terms of accuracy, demonstrating the potential of LLMs in effectively addressing partially observed signals in graphs.
\end{abstract}

\section{Introduction}

The application of Large Language Models (LLMs) in graph data processing and graph learning tasks has increasingly garnered research attention. GraphLLM, a graph learning model that integrates with large language models (LLMs), significantly enhances the ability of LLMs to reason with graph data by boosting accuracy and reducing the context length required for graph reasoning tasks \cite{chai2023graphllm}. The NLGraph benchmark is introduced to assess the graph reasoning abilities of language models, showing promising results on simple tasks but highlighting challenges with more complex problems, suggesting the need for further improvement \cite{wang2024can}.

In our work, we designed a framework that uses LLM to predict missing values in time-varying graph data by leveraging spatial and temporal smoothness. The LLM model aggregates signals from neighboring nodes and prior time steps, outperforming traditional methods in accuracy.

\begin{figure*}[htb]
\centering
\includegraphics[trim={10 10 0 15}, clip,width=0.95\textwidth]{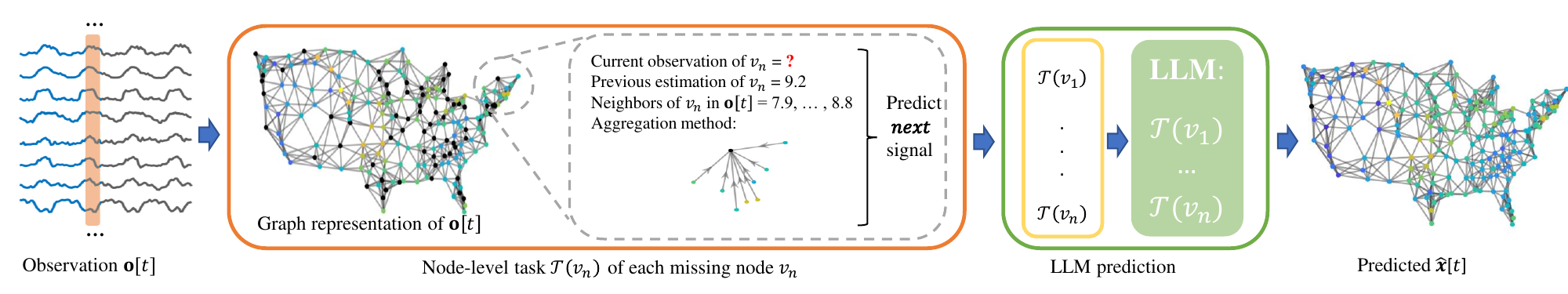}
\caption{An illustrative example of our algorithm}
\label{fig1}
\end{figure*}

\section{Methodology}
The proposed methodology leverages the smoothness assumption of data on graph nodes in both spatial and temporal dimensions on a time-varying regression task of online prediction. For a graph $\mathcal{G}$ with $N$ nodes, we assume that the time-varying signal (data value) on the graph nodes $\boldsymbol{x}[t]$ exhibits smoothness over time and space. This implies that for an observation $\boldsymbol{o}[t]$ containing missing node observations, the ground truth $\boldsymbol{x}[t]$ can potentially be inferred from the temporal trends or spatial proximity to other observed values. The Large Language Model (LLM) is trained to recognize and predict patterns that align with this smoothness assumption. In the context of graph-based formulation, a transductive setting naturally appears where observed data and unobserved data coexist, enabling the LLM to exploit the intrinsic structure of the graph for more accurate predictions on the unobserved nodes when the topological context is fed into LLM along with the data \cite{liu2023evaluating}.

To predict missing observations, the LLM is used as a message-passing mechanism realized by localized aggregation at each individual node:
\begin{equation}
    \text{agg}(x_v) = \Omega \left( \{ (x_v[t], x_u[t]) \mid u \in \mathcal{N}_v \} \right),
    \label{eq_message}
\end{equation}
where for the node $v$ in $\mathcal{G}$, $x_v[t]$ is the signal on $v$ and it has $\mathcal{N}_v$ 1-hop neighboring nodes (including self-loop) denoted with the subscript $u = 1  \dots N_v$.
We divide the global node observations $\boldsymbol{o}[t]$ and the estimation $\hat{\boldsymbol{x}}[t]$ into localized representation, where for each node $v$, we record only its own signal and its one-hop neighboring signals to feed into the context and the prompt of LLM.
At each time step $t$, the current partial observation ${o}_v[t]$ of a node, along with historical data (either observed or reconstructed), is provided as the input to the LLM. 
The LLM serves as a localized aggregation where its task $\mathcal{T}(v_n)$ is to interpret input from the neighbors of a missing node $v$ and the previous estimate of $\hat{x}_v[t-1]$, then inferring the missing observation $\hat{x}_v[t]$. Since this aggregation is defined locally on each node, the aggregation process is fed iterative to LLM for all $N$ nodes, which in turn is mapped back onto each missing node observation.
The algorithmic procedure is shown in Algorithm~\ref{alg:algorithm}.

\begin{algorithm}[htb]
\caption{Online Prediction of Unobserved Nodes}
\label{alg:algorithm}
\begin{algorithmic}[1]
\While{new observations $\boldsymbol{o}[t]$ are available}
\For{each missing node $v_n$ in  unlabelled node set}
\State Collect the previous estimation of $v_n$
\State Collect observed neighbor signals of $v_n$ in $\boldsymbol{o}[t]$
\State Form LLM task $\mathcal{T}(v_n)$ from aggregation \eqref{eq_message}
\State Feed $\mathcal{T}(v_n)$ into LLM and let LLM predict $\hat{{x}}_v[t]$
\EndFor
\State Collect all $\hat{{x}}_v[t]|_{n =1 ... M}$ and form $\hat{\boldsymbol{x}}[t]$
\EndWhile
\end{algorithmic}
\end{algorithm}

In Algorithm~\ref{alg:algorithm}, the process initiates by identifying any new observations at time $t$. For each missing node $v_n$ in $\mathcal{G}$, values from its own previous estimate and those of its neighbors in the current observations $\boldsymbol{o}[t]$ are collected. This collected data, including neighbor values from past estimations at $t-1$, informs the LLM task $\mathcal{T}(v_n)$, which then predicts the current value $\hat{x}_v[t]$ for the missing node. This prediction is iteratively performed for all missing nodes, culminating in a comprehensive state estimation $\hat{\boldsymbol{x}}[t]$ for time $t$.

By employing Algorithm~\ref{alg:algorithm}, the LLM iteratively learns the patterns from spatial neighbors and temporal trends to predict the missing data accurately. The approach also allows for the integration of LLM predictions with graph-based structures, enhancing the prediction process.

\section{Experiment Results and Discussion}
In our experiments, we employed time-varying hourly wind speed graph signals with $N = 197$ nodes and $T = 95$ time points to validate the proposed methodology from \cite{yan_2022_sign}. The data is uniformly set 30\% of the nodes to be missing. The GPT-3.5-turbo model from the OpenAI API was utilized to predict these missing values based on the temporal and spatial patterns in the data. 
GPT-3.5-turbo is an advanced LLM, part of the GPT-3 family, that excels in understanding and generating human-like text. It can handle complex prompts and generate highly accurate predictions, making it suitable for tasks involving pattern recognition and missing data prediction.

The GPT-3.5-turbo model was provided with the observations from the previous node to predict the missing values for the same node at the next time point. By inputting only one time point at a time, we can achieve online prediction and prevent the LLM from peeking into future data. Additionally, in the prompt, we explicitly instruct it not to use chat memory. Each algorithm is repeated 5 times. Then we can calculate the $\text{MSE} = \frac{1}{5NT} \sum_{i=1}^{N}\sum_{t=1}^{T} \left(x_i[t] - \hat{x}_i[t] \right)^2$ of the predicted value and evaluate the results. In the experiments, we used GLMS \cite{bib_LMS} and G-Sign \cite{yan_2022_sign} as baseline GSP algorithms for comparison. The result is shown as follows:

\captionsetup{font=small} 

\setlength{\tabcolsep}{25pt}

\begin{table}[ht]
\centering
\begin{tabular}{c c}
\toprule
\textbf{Model} & \textbf{MSE} \\
\midrule
GLMS Algorithm & 3.396 \\
G-Sign Algorithm & 3.718 \\
GPT-3.5-turbo & \textbf{1.560} \\
\bottomrule
\end{tabular}
\caption{MSE Comparison of Different Algorithms}
\end{table}

Based on the MSE comparison between different algorithms, the results demonstrate that the GPT-3.5-turbo model significantly outperforms both the GLMS and G-Sign algorithms in handling missing data within graph structures. This may be attributed to the fact that adaptive filters are primarily designed for noise reduction and are less effective for pure noiseless spatiotemporal prediction tasks compared to powerful models like GPT, which suggests that the GPT model holds substantial potential in addressing challenges related to partially observed time-varying graph signals. 

\noindent{\bf{Limitations}} There are rare occasions (occurred once during our 5 experiment runs) where the model outputs $NaN$ values and the cause remains unclear. In such cases, a practical workaround is employed: the missing values are substituted with the average of the previous time points. Additionally, batch-feeding $N$ nodes tasks into LLM often results in the number of returns not being $N$. While the GPT model demonstrates substantial potential, addressing these limitations through complementary strategies remains necessary to further improve its robustness in scenarios involving noisy, time-varying, and incomplete data.

\bibliography{main}

\end{document}